\documentclass[conference]{IEEEtran}
\usepackage[OT1]{fontenc}
\usepackage{cite}
\usepackage{amsmath,amssymb,amsfonts}
\usepackage{graphicx}
\usepackage{textcomp}
\usepackage{xcolor}
\usepackage{mathtools}
\usepackage{algorithm}
\usepackage{algpseudocode}
\usepackage{hyperref}

\def\BibTeX{{\rm B\kern-.05em{\sc i\kern-.025em b}\kern-.08em
    T\kern-.1667em\lower.7ex\hbox{E}\kern-.125emX}}

\begin{document}

\title{Multiscale Dubuc: A New Similarity Measure for Time Series}

\author{
    \IEEEauthorblockN{Mahsa~Khazaei\IEEEauthorrefmark{1},
    Azim~Ahmadzadeh\IEEEauthorrefmark{1},
    Krishna~Rukmini Puthucode\IEEEauthorrefmark{2}}
    \IEEEauthorblockA{
        \IEEEauthorrefmark{1}Department of Computer Science,
        University of Missouri-St. Louis,
        St. Louis, MO, USA\\
        \IEEEauthorrefmark{2}Department of Computer Science,
        Georgia State University,
        Atlanta, GA, USA\\
        Email: \IEEEauthorrefmark{1}mkdgy@umsystem.edu \\
        \vspace{-1.1cm}
    }
}

\date{September 2024}

\maketitle

\begin{abstract}
    Quantifying similarities between time series in a meaningful way remains a challenge in time series analysis, despite many advances in the field. Most real-world solutions still rely on a few popular measures, such as Euclidean Distance (EuD), Longest Common Subsequence (LCSS), and Dynamic Time Warping (DTW). The strengths and weaknesses of these measures have been studied extensively, and incremental improvements have been proposed. In this study, however, we present a different similarity measure that fuses the notion of Dubuc's variation from fractal analysis with the Intersection-over-Union (IoU) measure which is widely used in object recognition (also known as the Jaccard Index). In this proof-of-concept paper, we introduce the Multiscale Dubuc Distance (MDD) measure and prove that it is a metric, possessing desirable properties such as the triangle inequality. We use 95 datasets from the UCR Time Series Classification Archive to compare MDD's performance with EuD, LCSS, and DTW. Our experiments show that MDD's overall success, without any case-specific customization, is comparable to DTW with optimized window sizes per dataset. We also highlight several datasets where MDD's performance improves significantly when its single parameter is customized. This customization serves as a powerful tool for gauging MDD's sensitivity to noise. Lastly, we show that MDD's running time is linear in the length of the time series, which is crucial for real-world applications involving very large datasets.
    \vspace{-0.1cm}
\end{abstract}


\vspace{-0.1cm}
\section{Introduction}
    Time series data is a real-valued sequence that shows the progression of a random variable in time. Analyzing time series is relevant in many industries and applications, such as the financial market, health care, and space-weather forecasting. Finding similarity between time series is challenging due to differences in the definition of similarity in different applications. Consequently, many similarity measures have been proposed that work well in some tasks and under-perform in others, suggesting that no measure beats all other measures on all datasets.
    
    Similarity measures for time series fall into two main categories: Lock-step measures and elastic measures. Lock-step measures, like Euclidean distance, measure the similarity between two time series by mapping the \textit{i}-th element of the first time series with the \textit{i}-th element of the second time series. Euclidean distance \cite{faloutsos_fast_1994} is one of the most commonly used lock-step measures of similarity for time series. However, the two time series compared with this measure must be of the same length.
    
    The most commonly used elastic measures are Edit Distance on Real Sequence (EDR) \cite{chen2005robust}, Longest Common Subsequence (LCSS) \cite{vlachos2002discovering} and, Dynamic Time Warping (DTW) \cite{sakoe1987dynamic}. Measures in this group perform a one-to-many mapping of the time series. DTW is the most popular measure in this group, however, it is susceptible to pathological alignment, meaning that a single point can be mapped to a larger subset of points in the other time series \cite{sakoe1987dynamic}.
    
    Another key challenge in time series analysis is the need for a measure with a configurable level of sensitivity to detail. These fine-grained details are essential for defining similarity in some applications but could be safely ignored in others. Therefore, a measure with a configurable level of sensitivity to detail could ensure that domain experts can easily fine-tune the granularity levels to their liking.
    
    Recently, representation learning of time series has gained attention in light of current advancements in deep learning. In these works, time series, which are already high-dimensional, are transformed to even higher dimensions with the goal that by increasing the dimensionality, more information can be learned using Deep Neural Network architectures from the original time series. One challenge in this scenario is ensuring that the time series' similarity is preserved in latent space. It has been shown that some of these methods lack this fundamental feature and, thus, should be used cautiously \cite{foumani_series2vec_2024}. 

    This paper proposes Multiscale Dubuc Similarity ($\mathcal{MDS}$), a new similarity measure inspired by fractal analysis of time series and Dubuc's variation method. This new measure utilizes the multiscale characteristics of time series to quantify and compare the complexity of a pair of time series with adjustable sensitivity to fine-grained details. We show that this similarity measure and its corresponding distance measure ($\mathcal{MDD}$) can be used as a similarity/distance metric in practice as they satisfy the triangle inequality axiom.

\vspace{-0.1cm}
\section{Background}
    Many similarity measures have been proposed for time series; however, in this study, we have limited our review to measures most commonly used in the literature and those that are in some way comparable to our proposed measure.
    \vspace{-0.1cm}
    \subsection{State-of-the-Art Similarity Measures for Time Series}
        
        Euclidean distance is the square root of the sum of the squared differences between corresponding points in two time series data. In terms of time complexity and overall performance, it is a powerful metric but has several disadvantages. The first one is that it is a lock-step measure, so the mapping between the time series is not flexible as it is one-to-one. Also, to be comparable, time series should have the same length. Calculating Euclidean distance is computationally inexpensive. However, for applications that require dynamic mapping of time series and adjustable sensitivity to noise, time complexity is not necessarily a factor to be considered. This makes Euclidean distance less than ideal for those applications.

        \begin{figure}[t]
            \centering
            \includegraphics[width=\columnwidth]{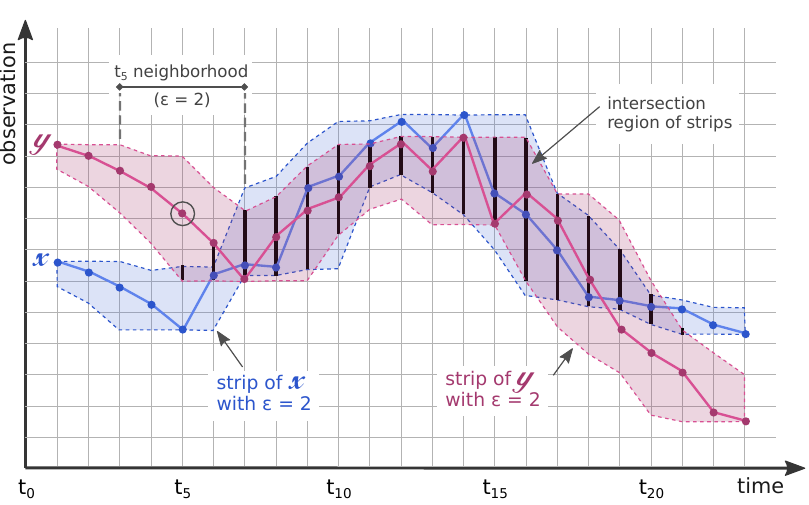}
            \caption{The graphic showing the idea behind the \textit{Multisclae Dubuc Similarity} ($\mathcal{MDS}$) measure. Two time series $(x_t)_{t=1}^{d}$ and $(y_t)_{t=1}^{d}$ are compared using their identified envelopes (regions with dashed borders) for $\varepsilon = 2$. $\mathcal{MDS}$ quantifies the similarity between the two time series at different granularity levels by changing $\varepsilon$, and computes the intersection ratio (see Eq.~\ref{eq:intersection-ratio}) at each step. It then aggregates the intersection ratios to return a real value between 0 and 1 as the similarity (or distance) between the two time series. The thick black bars represent the total intersection between the envelopes.}
            \label{fig:main_graphic}
            \vspace{-0.6cm}
        \end{figure}

        DTW is one of the most commonly used similarity measures due to minimizing the effects of distortion and noise. The DTW algorithm calculates the pairwise distance matrix between all elements of the two time series using Euclidean distance as its distance function. It then determines the shortest warping path between the two time series using the Bellman-Ford recursion. DTW is an effective measure as it allows for stretching or compressing the time axis to achieve a better match between time series. However, as previously discussed, this method is susceptible to pathological alignment. Variations of DTW have been proposed to address this problem by aligning based on the shape of the time series. One such variant is DDTW, which improves standard DTW by using the first derivatives of the time series to capture the local trends (i.e., rising, falling) and, instead of y-axis values, performs the alignment on these derivative sequences. This leads to more meaningful alignments in many cases by matching points that exhibit opposing local trends\cite{keogh_derivative_2001}. Similarly, shapeDTW aligns points using DTW based on their local descriptors (e.g., slope, Discrete Wavelet Transform, or Piecewise Aggregate Approximation) that capture the shape of the time series around a fixed-width neighborhood of that point \cite{zhao_shapedtw_2018}. This method avoids matching dissimilar regions, which often leads to pathological warpings in DTW.
       
        Due to not being differentiable, the original DTW can not be utilized in optimization tasks. SoftDTW is a differentiable version of DTW and can serve as a loss function in neural networks \cite{Cuturi2017softdtw}.

         Another category of similarity includes shape-based measures. 
         Kim et al. proposed a similarity measure for shape-based retrieval that, for every query, returns the most similar subsequence in a sequence database \cite{kim_shape-based_2002}. They use moving averages to compensate for noise and time warping to calculate the distance between the query and the subsequence. Another shape-based measure is SpADe \cite{chen_spade_2007}, a warping distance metric designed to handle shifting and scaling in temporal and amplitude of time series using local pattern matching, which existing metrics like Euclidean distance, DTW, LCSS, and EDR struggle handling. SpADe takes many parameters as input that need to be fine-tuned manually, namely, temporal and amplitude scales, the number of local patterns, the largest allowed gap, the penalty function on the gaps between local patterns, and the coefficient of the penalty function. 
        
        Some time series analysis approaches do not directly work on the original time series data, instead, they utilize a representation of them in a higher dimension. These methods rely on the concept of \textit{contrastive learning}, which uses positive and negative samples to train the model by minimizing the distance between the output and a positive sample, and maximizing its difference with a negative sample. This difference is then used to optimize a loss function, and for that, measuring the similarity between the learned representation and positive and negative samples is needed. The similarity of time series is defined differently in different applications, which makes finding positive and negative samples that are truly similar (or dissimilar) to the reference input a challenge of this method. 
        An example of such approaches is T-Loss \cite{franceschi_unsupervised_2019} where a triplet loss function is used to push the representation of a reference time series segment to be close to the positive examples and far from negative examples.

        Other studies have explored augmentation techniques for generating positive and negative pairs from a reference point. In TS2Vec\cite{yue_ts2vec_2022}, positive and negative pairs are generated by applying timestamp masking and random cropping to create two augmented contexts of the same sub-series. A softmax loss function is then used to optimize the representation on both the instance-level and temporal level.

        Foumani et al. argued that the works previously mentioned, under certain conditions, might produce negative samples that are more similar to the reference than the positive samples, which will introduce bias in downstream tasks \cite{foumani_series2vec_2024}.
        To this end, they proposed Series2Vec, a vector representation of time series using order-invariant attention mechanism and self-supervised representation learning. Their proposed method claims to preserve similarity, i.e., if two time series are similar in the original feature space, they will be similar in the learned representation.  
        This technique eliminates the data augmentation step, which is claimed to contribute to its similarity-preserving feature. 
        In this work, $\ell_1$ loss is used to maximize the similarity of the learned vectors in latent space (computed using the dot product of the vectors) and the similarity of raw time series data (computed using SoftDTW).

        It can be seen that more modern approaches focus on permutation-invariant models for encoding time series, while still relying on classical loss functions such as softmax and $\ell_1$ for optimizing their representation. Our approach, on the other hand, measures the similarity of time series in their original dimension ensuring that we are capturing the multi-resolution features of time series while providing explainability.

    \vspace{-0.1cm}
    \subsection{Comparable Ideas}
        The idea of capturing information from time series at different granularity levels has been long explored in the literature. Wavelet transform provides a way to analyze the microscopic trends present in the time series using a hierarchical framework. A wavelet transform is the inner product of the time series translated wavelet which is usually the \textit{n}-th derivative of a smoothing kernel \cite{struzik_haar_1999}. 
       
         More recently, this important feature of time series has been gaining attention. In TS2Vec (mentioned earlier), one of the motivations of the proposed framework is paying attention to the multi-resolution feature of time series by hierarchically discriminating positive and negative samples at different instance-wise and temporal scales to capture fine-grained details of the data. Another recent study that proposed a measure that uses this multi-resolution feature is TS-MIoU \cite{Ahmadzadeh2023tsmiou}, which is a time series similarity metric inspired by the same intuition as the works previously mentioned. The authors proposed that for certain tasks, finding similarity between two objects and two time series can be tackled using the same techniques, namely, \textit{box-counting} method, used to calculate the dimension of a fractal. However, this only applies if the box-sizes are proportional to the space in which the time series is defined.

\vspace{-0.1cm}
\section{Multiscale Dubuc}
    Fractal dimensions allow us to measure the intricacy of self-similar objects referred to as \textit{fractals} \cite{mandelbrot1982fractal}. There are different methods to calculate fractal dimensions, such as the similarity dimension and the box-counting dimension \cite{theiler1990estimating}. Intuitively, time series can be viewed as self-similar ``shapes'' even though they lack the spatial dimension seen in fractals. So, caution must be taken when using methods like box-counting to quantify the dimension of time series. The primary concern in this case is the undefinable spatial aspect ratio (between time and observation axes) of time series. To address this issue, alternative methods have been proposed that adapt the box-counting technique to the one-dimensional nature of time series data. In this section, we will first review Multiscale Intersection-over-Union for time series (TS-MIoU), one recent study that was particularly motivated by fractal analysis of time series and how they handled the undefined aspect ratio. Then, we will discuss another interesting approach that lets us use fractal analysis tools without defining the aspect ratio.

    \vspace{-0.1cm}
    \subsection{TS-MIoU}
        TS-MIoU is a region-based time series similarity measure based on an idea originally proposed as an object detection evaluation metric called Multiscale Intersection-over-Union (MIoU) \cite{ahmadzadeh2021miou}. MIoU is a fusion between Intersection-over-Union (IoU), measuring the relative overlap between two regions, and fractal dimension, quantifying the geometric complexity of fractal-like shapes. MIoU claims to improve IoU's sensitivity to structural details across multiple resolutions using the box-counting method. This multiscale approach allows for identifying misalignments at different granularity levels. The idea behind TS-MIoU is to treat time series as objects and apply grid-based segmentation to them. However, treating time series as objects results in an ill-defined space that was addressed in the original study by using non-square grid cells for segmentation, referred to as \textit{Segmentation with Proportional Binning}. This solution ensures that the boxes scale proportionally with the time series data.

        Since we will not be treating time series as an object, the new metric presented in this study does not suffer from the ill-definedness of space issue.

    \vspace{-0.2cm}
    \subsection{Dubuc's Variation Method}
        As mentioned earlier, applying the box-counting method to time series, typically used for computing the fractal dimension of shapes, results in an ill-defined space. One way to utilize the fractal features of time series without having to deal with the arbitrariness of the space is to use a method called \textit{Variational Box-Counting} \cite{taylor2019fractal}. In this method, instead of superimposing a fixed grid, boxes can be moved vertically to allow a better fix on the data. This method improves on the traditional box-counting method by reducing the number of boxes needed to cover the figure in question \cite{taylor2019fractal}.
        
        Dubuc's Variation method improves on the variational box-counting method by avoiding the treatment of time series as a spatial figure, thus not exhibiting the ill-definedness of space problem while utilizing the multi-scale characteristics of time series. This approach quantifies the fractal dimension of a time series by using the space-filling characteristics of its trace and measuring its amplitude at different time scales using a sliding window approach \cite{dubuc1989evaluating, taylor2019fractal}. For a given time series $(x_t)_{t=1}^{d}$ of length $d$, to measure the amplitude of the trace, the \textit{upper bound} ($u_{x, \varepsilon}(t)$) and \textit{lower bound} ($l_{x, \varepsilon}(t)$) of the trace at a given timestamp $t$ are defined as $u_{x, \varepsilon}(t) = \sup_{t' \in R_{\varepsilon}(t)} x_{t'} ,  l_{x, \varepsilon}(t) = \inf_{t' \in R_{\varepsilon}(t)} x_{t'}$ where the neighborhood $R_{\varepsilon}(t)$ is defined as:

        \vspace{-0.4cm}
        \begin{equation}
            R_{\varepsilon}(t) = \{s : |t - s| \leq \varepsilon \text{ and } s \in [1, d]\}.
        \end{equation}

        The $\varepsilon$ variation of the time series used in $R_{\varepsilon}(t)$ is then computed as follows:
        
        \vspace{-0.7cm}
        \begin{equation}
            V(\varepsilon) = \frac{1}{\varepsilon ^ 2}\sum_{t=1}^{d} \bigl( u_{\varepsilon}(t) - l_{\varepsilon}(t) \bigr).
        \end{equation}

        Essentially, Dubuc's variation method computes an envelope around the trace of the time series, and the width of this envelope is controlled by $\varepsilon$. For $\varepsilon = 0$ the envelope is equivalent to the trace, and the larger the $\varepsilon$ gets, the thicker the envelope becomes, focusing on macroscopic trends in the time series. We will leverage this multi-scale feature of Dubuc's variation method to quantify the similarity of two time series.

\vspace{-0.1cm}        
\section{Multiscale Dubuc Similarity Measure}
    We take advantage of the envelope defined by Dubuc's variation method to construct a similarity measure. Given two time series, $(x_t)_{t}^{d}$ and $(y_t)_{t}^{d}$, the total intersection and union of their corresponding envelopes can be defined as follows:

    \vspace{-0.6cm}
    \begin{align}
        \begin{split}
            \cap_{\varepsilon}(x, y) &= \sum_{t = 1}^{d} \max \bigl( \min(u_{x, \varepsilon}(t), u_{y, \varepsilon}(t)) \\
            &\quad\quad\quad\quad\quad - \max(l_{x, \varepsilon}(t), l_{y, \varepsilon}(t)), 0 \bigr), \\
            \cup_{\varepsilon}(x, y) &= \sum_{t = 1}^{d} \max \bigl( \max(u_{x, \varepsilon}(t), u_{y, \varepsilon}(t)) \\
            &\quad\quad\quad\quad\quad - \min(l_{x, \varepsilon}(t), l_{y, \varepsilon}(t)), 0 \bigr).
        \end{split}\label{eq:union-and-inters}
    \end{align}

    \noindent Note that the notion of ``intersection'' and ``union'' as defined in Eq.~\ref{eq:union-and-inters} should not be equated with the common understanding of those concepts when used on shapes. Our definition only takes into account the observation values of the two time series, because time series are sequences and not continuous functions. Our notion of intersection is illustrated in Fig.~\ref{fig:main_graphic}.

    The union in Eq.~\ref{eq:union-and-inters} is needed for normalizing the intersection value and transforming it into the range $[0, 1]$, which gives us the \textit{intersection ratio} as follows:
    
    \vspace{-0.4cm}
    \begin{equation}
        r(x,y,\varepsilon) = \frac{\cap_{\varepsilon}(x, y)}{\cup_{\varepsilon}(x, y)}.
        \label{eq:intersection-ratio}
    \end{equation}

    For a pair of time series $x$ and $y$, and a set $\mathcal{E}$ containing the candidate values for $\varepsilon$, \textit{Multiscale Dubuc Similarity} ($\mathcal{MDS}$) measure can then be defined as the area under the curve of the intersection ratios:

    \vspace{-0.6cm}
    \begin{equation}
        \mathcal{MDS}(x,y, \mathcal{E}) = \sum_{i=2}^{\mathcal{|E|}} \frac{r(x, y, \varepsilon_{i-1}) + r(x, y, \varepsilon_i)}{2} \Delta\varepsilon_i,
        \label{eq:trapezoidal}
    \end{equation}
    \vspace{-0.5cm}

    \noindent where $\Delta\varepsilon_i = \varepsilon_i - \varepsilon_{i-1}$.

    \setlength{\textfloatsep}{0pt}
    \begin{algorithm}[t]
        \caption{Multiscale Dubuc as a distance metric}\label{alg:pseudocode}
        \begin{algorithmic}
        \Function{MultiscaleDubuc}{$x, y, \mathcal{E}$}
        \State \textbf{initialize} ratios $\gets \{\}$
    
        \For{$\varepsilon$ in $\mathcal{E}$}
            \State $s_1 \gets \text{calculate-bounds}(x, \varepsilon)$
            \State $s_2 \gets \text{calculate-bounds}(y, \varepsilon)$
            \State $ \cap_{\varepsilon} \gets \text{intersection}(s_1, s_2)$
            \State $ \cup_{\varepsilon} \gets \text{union}(s_1, s_2)$
            \State $r \gets I / U$
            \State \text{ratios.append}(r)
        \EndFor
    
        \State $\mathcal{MDD}$ $\gets 1 - \text{area-under-the-curve}(\text{ratios})$
        \State \Return $\mathcal{MDD}$
        \EndFunction
        \end{algorithmic}
    \end{algorithm}
    
    \vspace{-0.1cm}
    \subsection{Multiscale Dubuc Distance as a Metric}
        The \textit{Mulstiscale Dubuc Distance}($\mathcal{MDD}$), defined as $\mathcal{MDD}(x, y, \varepsilon) = 1 - \mathcal{MDS}(x, y, \varepsilon)$, is a metric. More accurately, it's a pseudometric considering very rare cases. To show this, we review the conditions that any metric must hold.

        Let $\mathcal{X}$ be a set and $d: \mathcal{X} \times \mathcal{X} \longrightarrow \mathbb{R}$ be a distance function. We say $d$ is a metric on $\mathcal{X}$ if the following conditions are held for all $x,y,z \in \mathcal{X}$: (1) \textit{positiveness}, $d(x,y)\geq 0$; (2) \textit{strict positiveness}, $x \neq y \Rightarrow d(x,y)>0$; (3) \textit{symmetry}, $d(x,y)=d(y,x)$; (4) \textit{reflexivity}, $d(x,x)=0$; and (5) \textit{triangle inequality}, $d(x,z) \leq d(x,y) + d(y,z)$. A pseudometric is a metric that relaxes the strict positiveness condition, i.e., in some rare cases, it may allow the distance between two distinct time series to be zero.

        For any given pair of time series $x, y \in \mathcal{X}$, both $\cup_\varepsilon$ and $\cap_\varepsilon$ operations are always non-negative because $\max(\_,0)\ge0$, which guarantees positiveness of $\mathcal{MDD}$. The strict positiveness condition is true if, from \textit{all} pairs of envelopes obtained from two distinct time series, at least one pair consists of distinct envelopes (i.e., $\exists \varepsilon \in \mathcal{E}\; s.t. \cap_\varepsilon(x,y) \neq \cup_\varepsilon(x,y)$). This cannot be mathematically guaranteed as it depends on $\mathcal{E}$ and the dataset itself, however, it is extremely unlikely that for all values of $\varepsilon \in \mathcal{E}$, all of the corresponding envelopes are pair-wise identical. $\mathcal{MDD}$ inherits the symmetry property from $r$ (i.e., $r(x, y, \varepsilon) = r(y, x, \varepsilon))$, which in turn, inherits it from $\cap_\varepsilon$ and $\cup_\varepsilon$;  $\forall x,y \in \mathcal{X}, \cap_\varepsilon(x, y) = \cap_\varepsilon(y,x)$ and $\cup_\varepsilon(x,y) = \cup_\varepsilon(y, x)$). The reflexivity condition is also held: for any $x \in \mathcal{X},\;\;$ $\cap_\varepsilon(x, x) = \cup_\varepsilon(x,x)$ results in $r(x, y, \varepsilon) = 1$ for any $\varepsilon \in \mathcal{E}$, and consequently, $\mathcal{MDD}(x, x, \varepsilon) = 0$. Lastly, $\mathcal{MDD}$ inherits the triangle inequality condition from the intersection ratio $r$. This was shown in detail for the Jaccard Index, which is identical to how we defined $r$ \cite{gilbert1972distance, kosub2019note}.
        
    \vspace{-0.1cm}
    \subsection{Time Complexity of Multiscale Dubuc Distance}
        A pseudocode of $\mathcal{MDD}$’s algorithm is given in Alg.~\ref{alg:pseudocode}. The `calculate-bounds' method calculates the upper and lower bounds of a pair of time series by iterating through the time series once, hence $\Theta(d)$. The intersection and union operations each take linear time as Eq.~\ref{eq:union-and-inters} indicates. The `area-under-the-curve' method runs Eq.~\ref{eq:trapezoidal} which is linear in $|\mathcal{E}|$. This brings the total running time of $\mathcal{MDD}$’s algorithm to $\Theta(|\mathcal{E}|\cdot d)$. However, $\mathcal{E}$ is a fixed set of values the user defines and does not scale with the problem size. Therefore, the running time of this algorithm is $\Theta(d)$, i.e., linear in the length of time series.


\vspace{-0.1cm}
\section{Experiments and Results}\label{sec:experiments-and-results}
    \vspace{-0.1cm}
    \subsection{Experimental Settings}
        In this study we test our novel distance metric on 95 datasets (out of 128) from the UCR Time series Archive \cite{UCRArchive2018}
        \footnote{The source code is available at \\
        \href{https://bitbucket.org/dataresearchlab/multiscale_dubuc}{https://bitbucket.org/dataresearchlab/multiscale\_dubuc}}. We excluded 11 datasets with time series of varying lengths, as different similarity measures handle varying lengths differently, potentially confounding our experiments. Additionally, we excluded 22 datasets with lengthy time series (more than 900 observations), which was based on the understanding that analyzing full-length time series of such magnitude has limited real-world applications. The performance of a similarity measure on these lengthy time series does not provide substantial insight into its strengths and weaknesses.
        
        We compared $\mathcal{MDD}$ with Euclidean distance (EuD), DTW, and LCSS. Before conducting the experiment, each dataset's upper and lower bounds were computed. Note that $\mathcal{MDD}$ depends on how $\mathcal{E}$ is defined; larger values of $\varepsilon \in \mathcal{E}$ result in a lower sensitivity to noise, whereas smaller values of $\varepsilon$ render $\mathcal{MDD}$ more sensitive to noise. In our experiments, we determined $\mathcal{E}$ generically. That is, we used powers of two between 1 and a fraction ($\alpha = 0.4$) of the length of the time series. For example, for $\alpha=0.4$ and a dataset of time series of length 200, we determined $\mathcal{E}$ to be $\{1, 2, 4, \cdots, 64\}$, i.e., all powers of two between 1 and $0.4 * 200$.

    \vspace{-0.1cm}
    \subsection{Accuracy Gain of Multiscale Dubuc}
        To test the performance of $\mathcal{MDD}$, we followed the evaluation framework recommended in \cite{batista2011complexity, dau2019ucr}. We computed the accuracy of a 1-NN classifier by performing leave-one-out cross-validation, using $\mathcal{MDD}$ as its distance metric. The result of this experiment on the training set of each dataset gave us the \textit{expected} accuracy and is denoted as $\widehat{\text{acc}}_{\text{$\mathcal{MDD}$}}$. We then ran 1-NN on the test set of each dataset. During testing, each instance's label was classified based on its most similar instance in the training set. This gave us the \textit{actual} accuracy obtained by $\mathcal{MDD}$ on a given dataset, denoted as $\text{acc}_{\text{$\mathcal{MDD}$}}$.
        
        To obtain a relative performance gain on each dataset, for an arbitrary distance measure $\mu$ ($\mathcal{MDD}$ here), the \textit{expected accuracy gain} and \textit{actual accuracy gain} with respect to a reference distance measure $\mu_{ref}$ (i.e., DTW, EuD, LCSS, and TS-MIoU) is computed as $\widehat{g}_{\text{$\mu$,ref}} = \frac{\widehat{\text{acc}}_{\mu}}{\widehat{\text{acc}}_{\mu_{ref}}}$ and $g_{\text{$\mu$, ref}} = \frac{\text{acc}_{\mu}}{\text{acc}_{\mu_{ref}}}$, respectively.
        We used the ``Texas Sharpshooter plot" for visualizing the accuracy gain of $\mathcal{MDD}$ over the other measures \cite{dau2019ucr}. The four sections of this plot, as visualized in Fig.~\ref{fig:sharpshooter_plot}, are true positive (TP), false positive (FP), true negative (TN), and false negative (FN).

        As for the baseline accuracy of 1-NN using other distance measures, for DTW, we used the accuracy of DTW with learned window size as reported on the UCR's website \cite{UCRArchive2018}. For other measures we reran the experiment described above and used the $tslearn$ Python package for implementation of LCSS \cite{tavenard2020tslearn}. For TS-MIoU, we used the source code made publicly available by the authors\cite{code-tsmiou}.

    \vspace{-0.1cm}
     \subsection{Results and Discussions}
\begin{figure}[t]
            \centering
            \includegraphics[width=\columnwidth]{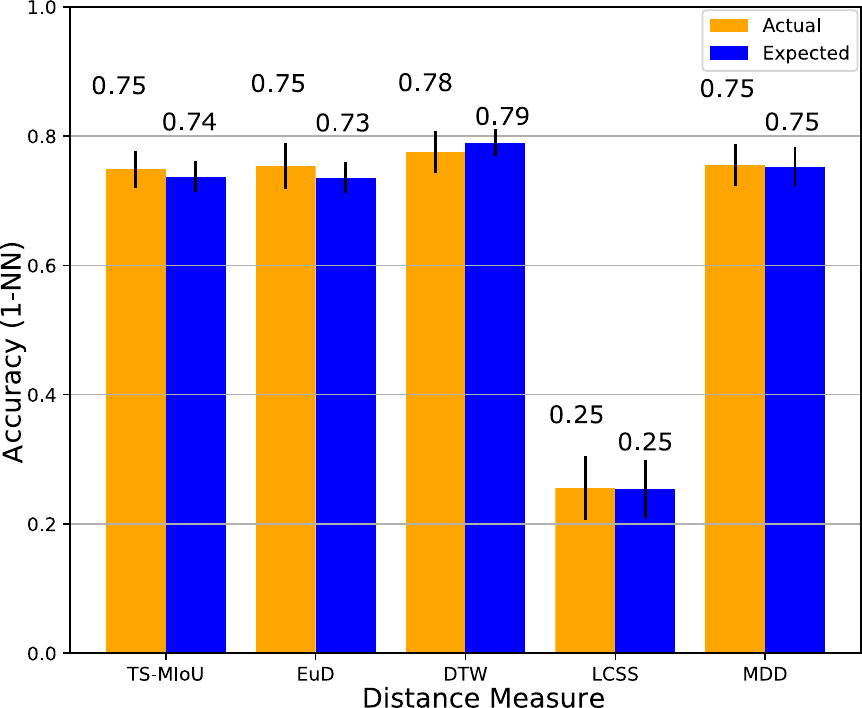}
            \caption{The bar plot showing the average expected (blue) and actual (orange) accuracy values of the 1-NN classifier on 95 UCR datasets, using TS-MIoU, EuD, DTW, and $\mathcal{MDD}$ distance functions.}
            \label{fig:barplot_accuracy}
            \vspace{-0.3cm}
        \end{figure}

        Fig.~\ref{fig:barplot_accuracy} shows the average accuracy obtained from each distance metric. The aggregated results illustrated in Fig.~\ref{fig:barplot_accuracy} show that $\mathcal{MDD}$ (with $\text{acc}_{\text{$\mathcal{MDD}$}}=0.75 \pm 0.03$, $\widehat{\text{acc}}_{\text{$\mathcal{MDD}$}}=0.75 \pm 0.02$) can indeed make 1-NN classifier to achieve a performance similar to
        DTW (with $\text{acc}_{\text{DTW}}=0.78 \pm 0.03$, $\widehat{\text{acc}}_{\text{DTW}}=0.79 \pm 0.02$), TS-MIoU (with $\text{acc}_{\text{TS-MIoU}}=0.75 \pm 0.02$, $\widehat{\text{acc}}_{\text{TS-MIoU}}=0.74 \pm 0.02$) and EuD (with $\text{acc}_{\text{EuD}}=0.75 \pm 0.03$, $\widehat{\text{acc}}_{\text{EuD}}=0.73 \pm 0.02$). Note that in this comparison, DTW's parameter, the \textit{warping window size}, was already optimized for each dataset, whereas for $\mathcal{MDD}$, a generic set of $\varepsilon$ values was used. 1-NN with LCSS performed significantly worse than the others, which resulted in an outside-the-range accuracy gain of $\mathcal{MDD}$, so we removed its corresponding points from the Texas Sharpshooter plot for better visibility.
        
\begin{figure}[t]
            \centering
            \includegraphics[width=\columnwidth]{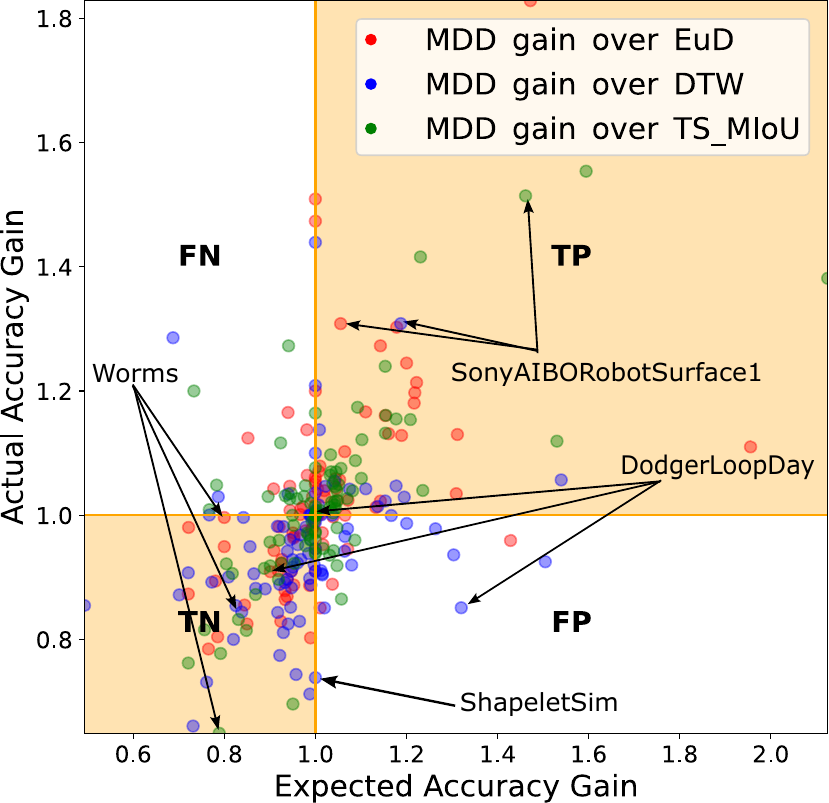}
            \caption{The Texas Sharpshooter plot showing the accuracy gain of $\mathcal{MDD}$ over EuD, DTW, and TS-MIoU.}
            \label{fig:sharpshooter_plot}
            \vspace{0.1cm}
        \end{figure}

        As shown in Fig.~\ref{fig:sharpshooter_plot}, $\mathcal{MDD}$ results in an accuracy greater than or equal to that of EuD, DTW with learned parameter, LCSS, and TS-MIoU on $56\%$, $31\%$, $95\%$, and $61\%$ of datasets, respectively. Focusing on the distribution of the points in Fig.~\ref{fig:sharpshooter_plot}, the further the points are from the center, the higher the accuracy gain is. If we look at all the points in the TP region that are far from the center, it becomes evident that the enhancement offered by our metric is not marginal, nor is it anecdotal. As an example, for the \verb|SonyAIBORobotSurface1| dataset, $\mathcal{MDD}$ improved 1-NN's performance by $51\%$, $30\%$ and $30\%$ compared to TS-MIoU, EuD, and DTW, respectively. 

        One of the main features of our metric is its customizable sensitivity to noise. For example, consider the \verb|Worms| dataset in Fig.~\ref{fig:sharpshooter_plot}. Initially, with the generic set of $\varepsilon$ values $\{1,2,4, \cdots,256\}$ the actual accuracy of $\mathcal{MDD}$ was $45\%$. However, once we took a closer look at the dataset, we noticed the impact of noise on 1-NN's performance. Our generic choice of $\varepsilon$ included values as small as $\varepsilon=1$, which results in high sensitivity to noise. Therefore, by excluding the smaller values we redefined $\mathcal{E}$ to be $\{16,32,...,256\}$. This decision boosted the accuracy of 1-NN to $63\%$, resulting in a $40\%$ improvement compared to the generic $\mathcal{E}$, $18\%$ and $40\%$ improvement compared to DTW and EuD respectively. This case study shows how $\mathcal{MDD}$ allows for gauging the amount of noise that should be tolerated by a classifier.
       
        It is important to note that $\mathcal{MDD}$ underperforms in certain cases compared to other measures. While we observed numerical improvements in several cases after customizing $\mathcal{E}$, the improvements were not always quantitatively significant. The \verb|ShapeletSim| dataset is one of such cases where even after customizing $\mathcal{E}$, the $g_{\text{$\mathcal{MDD}$, DTW}}$ and $g_{\text{$\mathcal{MDD}$, EuD}}$ remain in the FN region. This is one avenue we would like to further investigate as it is not clear to us what exactly renders $\mathcal{MDD}$ ineffective on certain datasets.

\vspace{-0.2cm}
\section{Conclusion, and Future Work}
    In this paper, we proposed Multiscale Dubuc Similarity $\mathcal{MDS}$, a new similarity metric for time series inspired by fractal time series analysis. Our metric provides a method of utilizing the multi-resolution characteristics of time series without dealing with an ill-defined space due to treating time series as spatial figures. We showed that the performance of our metric is comparable to state-of-the-art measures and that we provide performance gain in some datasets.
    Moreover, we proved that $\mathcal{MDD}$ is a metric satisfying the triangle inequality. 
    
    In this proof-of-concept paper, we only focused on comparing pair-wise elements of the time series, but one extension of this work can be comparing the envelopes corresponding to each time series instead of doing a pair-wise comparison.

\vspace{-0.2cm}
\section*{Acknowledgements}
    This material is based upon work supported by the National Science Foundation under Grant No. 2433781, directorate for Computer and Information Science and Engineering (CISE), and Office of Advanced Cyberinfrastructure (OAC).

\bibliographystyle{splncs04}
\raggedright
\vspace{-0.3cm}
\bibliography{main}
\end{document}